\documentclass[]{llncs}
\usepackage{url}
\usepackage{graphicx}
\usepackage{amsmath,amssymb} 
\usepackage{color}
\usepackage[width=122mm,left=12mm,paperwidth=146mm,height=193mm,top=12mm,paperheight=217mm]{geometry}
\usepackage[colorlinks,
            linkcolor=black,
            anchorcolor=black,
            citecolor=black,
            urlcolor=black
            ]{hyperref}
\begin{document}
\pagestyle{headings}
\mainmatter
\def\ECCV18SubNumber{1881}  

\title{A Systematic DNN Weight Pruning Framework using Alternating Direction Method of Multipliers} 

\author{Tianyun Zhang$^{1*}$, Shaokai Ye$^{1*}$, Kaiqi Zhang$^1$, Jian Tang$^1$, Wujie Wen$^2$, Makan Fardad$^1$ \& Yanzhi Wang$^3$\\
1. Syracuse University 2. Florida International University  \\ 3. Northeastern University  $^*$Equal Contribution\\
\texttt{1. {\{tzhan120,sye106,kzhang17,jtang02,makan\}@syr.edu}}\\
\texttt{2. wwen@fiu.edu} ~~ \texttt{3. yanz.wang@northeastern.edu}
}
\institute{}

\maketitle

\begin{abstract}
Weight pruning methods for deep neural networks (DNNs) have been investigated recently, but prior work in this area is mainly heuristic, iterative pruning, thereby lacking guarantees on the weight reduction ratio and convergence time. To mitigate these limitations, we present a systematic weight pruning framework of DNNs using the alternating direction method of multipliers (ADMM). We first formulate the weight pruning problem of DNNs as a nonconvex optimization problem with combinatorial constraints specifying the sparsity requirements, and then adopt the ADMM framework for systematic weight pruning. By using ADMM, the original nonconvex optimization problem is decomposed into two subproblems that are solved iteratively. One of these subproblems can be solved using stochastic gradient descent, the other can be solved analytically. Besides, our method achieves a fast convergence rate.

The weight pruning results are very promising and consistently outperform the prior work. On the LeNet-5 model for the MNIST data set, we achieve 71.2$\times$ weight reduction without accuracy loss. On the AlexNet model for the ImageNet data set, we achieve 21$\times$ weight reduction without accuracy loss. When we focus on the convolutional layer pruning for computation reductions, we can reduce the total computation by five times compared with the prior work (achieving a total of 13.4$\times$ weight reduction in convolutional layers). Our models and codes are released at \url{https://github.com/KaiqiZhang/admm-pruning}.

\end{abstract}

\section{Introduction}

Large-scale deep neural networks or DNNs have made breakthroughs in many fields, such as image recognition \cite{krizhevsky2012imagenet,lecun1998,he2016deep}, speech recognition \cite{hinton2012deep,dahl2012context}, game playing \cite{mnih2013playing}, and driver-less cars \cite{makantasis2015deep}. 
Despite the huge success, their large model size and computational requirements will add significant burden to state-of-the-art computing systems \cite{krizhevsky2012imagenet,simonyan2014,DeepCompression}, especially for embedded and IoT systems. As a result, a number of prior works are dedicated to \emph{model compression} in order to simultaneously reduce the computation and model storage requirements of DNNs, with minor effect on the overall accuracy. These model compression techniques include \emph{weight pruning} \cite{DeepCompression,han2015,dai2017,yang2016,molchanov2017variational,guo2016dynamic,tung2017fine,luo2017thinet}, sparsity regularization \cite{wen2016learning,liu2015sparse,zhou2016less}, weight clustering \cite{DeepCompression,chen2015compressing,park2017weighted}, and low rank approximation \cite{denil2013predicting,denton2014exploiting}, etc. 

A simple but effective weight pruning method has been proposed in \cite{han2015}, which prunes the relatively less important weights and performs retraining for maintaining accuracy in an iterative manner. It can achieve 9$\times$ weight reduction ratio on the AlexNet model with virtually no accuracy degradation. This method has been extended and generalized in multiple directions, including energy efficiency-aware pruning \cite{yang2016}, structure-preserved pruning using regularization methods \cite{wen2016learning}, and employing more powerful (and time-consuming) heuristics such as evolutionary algorithms \cite{dai2017}. 
While existing pruning methods achieve good model compression ratios, they are heuristic (and therefore cannot achieve optimal compression ratio), lack theoretical guarantees on compression performance, and require time-consuming iterative retraining processes.

To mitigate these shortcomings, we present a systematic framework of weight pruning and model compression, by (i) formulating the weight pruning problem as a constrained nonconvex optimization problem with combinatorial constraints, which employs the cardinality function to induce sparsity of the weights, and (ii) adopting the \emph{alternating direction method of multipliers} (ADMM) \cite{boyd2011} for systematically solving this optimization problem. By using ADMM, the original nonconvex optimization problem is decomposed into two subproblems that are solved iteratively. In the weight pruning problem, one of these subproblems can be solved using stochastic gradient descent, and the other can be solved analytically.
Upon convergence of ADMM, we remove the weights which are (close to) zero and retrain the network. 

Our extensive numerical experiments indicate that ADMM works very well in weight pruning.
The weight pruning results consistently outperform the prior work. On the LeNet-5 model for the MNIST data set, we achieve 71.2$\times$ weight reduction without accuracy loss, which is 5.9 times compared with \cite{han2015}. On the AlexNet model for the ImageNet data set, we achieve 21$\times$ weight reduction without accuracy loss, which is 2.3 times compared with \cite{han2015}. Moreover, when we focus on the convolutional layer pruning for computation reductions, we can reduce the total computation by five times compared with the prior work (achieving a total of 13.4$\times$ weight reduction in convolutional layers). Our models and codes are released at \url{https://github.com/KaiqiZhang/admm-pruning}.

\section{Related Work on Weight Reduction/Model Compression}

Mathematical investigations have demonstrated a significant margin for weight reduction in DNNs due to the redundancy across filters and channels, and a number of prior works leverage this property to reduce weight storage. The techniques can be classified into two categories:
\emph{1) Low rank approximation} methods \cite{denil2013predicting,denton2014exploiting} such as Singular Value Decomposition (SVD), which are typically difficult to achieve zero accuracy degradation with compression, especially for very large DNNs;
\emph{2) Weight pruning} methods which aim to remove the redundant or less important weights, thereby achieving model compression with negligible accuracy loss. 

A prior work \cite{han2015} serves as a pioneering work for weight pruning. It uses a heuristic method of iteratively pruning the unimportant weights (weights with small magnitudes) and retraining the DNN. It can achieve a good weight reduction ratio, e.g., 9$\times$ for AlexNet, with virtually zero accuracy degradation, and can be combined with other model compression techniques such as weight clustering \cite{DeepCompression,chen2015compressing}. It has been extended in several works. For instance, the \emph{energy efficiency-aware pruning} method \cite{yang2016} has been proposed to facilitate energy-efficient hardware implementations, allowing for certain accuracy degradation. The \emph{structured sparsity learning} technique has been proposed to partially overcome the limitation in \cite{han2015} of irregular network structure after pruning. However, neither technique can outperform the original method \cite{han2015} in terms of compression ratio under the same accuracy. There is recent work \cite{dai2017} that employs an evolutionary algorithm for weight pruning, which incorporates randomness in both pruning and growing of weights, following certain probabilistic rules. Despite the higher compression ratio it achieves, it suffers from a prohibitively long retraining phase. For example, it needs to start with an already-compressed model for further pruning on the ImageNet data set, instead of the original AlexNet or VGG models.

In summary, the prior weight pruning methods are highly heuristic and suffer from a long retraining phase. On the other hand, our proposed method is a systematic framework, achieves higher compression ratio, exhibits faster convergence rate, and is also general for structured pruning and weight clustering.

\section{Background of ADMM}
ADMM was first introduced in the 1970s, and theoretical results in the following decades have been collected in \cite{boyd2011}. It is a powerful method for solving regularized convex optimization problems, especially for problems in applied statistics and machine learning. Moreover, recent works \cite{takapoui2017simple,Leng2017} demonstrate that ADMM is also a good tool for solving nonconvex problems, potentially with combinatorial constraints, since it can converge to a solution that may not be globally optimal but is sufficiently good for many applications. 

For some problems which are difficult to solve directly, we can use variable splitting first, and then employ ADMM to decompose the problem into two subproblems that can be solved separately and efficiently. For example, the optimization problem 
\begin{equation}
\label{optz}
 \underset{\bf{x}}{\text{minimize}}
\ \ \ f({\bf{x}}) + g({\bf{x}}), 
\end{equation}
assumes that $f(\cdot)$ is differentiable and $g(\cdot)$ is non-differentiable but has exploitable structure properties. Common instances of $g$ are the $\ell_1$ norm and the indicator function of a constraint set. To make it suitable for the application of ADMM, we use variable splitting to rewrite the problem as 
\begin{equation*}
\begin{aligned}
& \underset{\bf{x, \ z}}{\text{minimize}}
& & f({\bf{x}}) + g({\bf{z}}), 
\\ & \text{subject to}
& & \bf{x} = \bf{z}.
\end{aligned}
\end{equation*}
Next, via the introduction of the augmented Lagrangian, the above optimization problem can be decomposed into two subproblems in $\bf{x}$ and $\bf{z}$ \cite{boyd2011}. The first subproblem is $\underset{\bf{x}} {\text{minimize}}\ f({\bf{x}})+q_1({\bf{x}})$, where $q_1(\cdot)$ is a quadratic function of its argument. Since $f$ and $q_1$ are differentiable, the first subproblem can be solved by gradient descent. The second subproblem is $\underset{\bf{z}} {\text{minimize}}\ g({\bf{z}})+q_2({\bf{z}})$, where $q_2(\cdot)$ is a quadratic function of its argument. In problems where $g$ has some special structure, for instance if it is a regularizer in (\ref{optz}), exploiting the properties of $g$ may allow this problem to be solved analytically. More details regarding the application of ADMM to the weight pruning problem will be demonstrated in Section 4.2. 

\section{Problem Formulation and Proposed Framework}

\subsection{Problem Formulation of Weight Pruning}

Consider an $N$-layer DNN, where the collection of weights in the $i$-th (convolutional or fully-connected) layer is denoted by ${\bf{W}}_{i}$ and the collection of biases in the $i$-th layer is denoted by ${\bf{b}}_{i}$. In a convolutional layer the weights are organized in a four-dimensional tensor and in a fully-connected layer they are organized in a two-dimensional matrix \cite{Leng2017}. 

Assume that the input to the (fully-connected) DNN is ${\bf{x}}$. Every column of ${\bf{x}}$ corresponds to a training image, and the number $t$ of columns determines the number of training images in the input batch. The input ${\bf{x}}$ will enter the first layer and the output of the first layer is calculated by
\begin{equation*}
{\bf{h}}_{1}=\sigma({\bf{W}}_{1}{\bf{x}}+{\bf{b}}_{1}),
\end{equation*}
where ${\bf{h}}_{1}$ and ${\bf{b}}_{1}$ have $t$ columns, and ${\bf{b}}_{1}$ is a matrix with identical columns. The non-linear activation function $\sigma(\cdot)$ acts entry-wise on its argument, and is typically chosen to be the ReLU function \cite{maas2013rectifier} in state-of-the-art DNNs.
Since the output of one layer is the input of the next, the output of the $i$-th layer for $i = 2, \ldots, N-1$ is given by
\begin{equation*}
{\bf{h}}_{i}=\sigma({\bf{W}}_{i}{\bf{h}}_{i-1}+{\bf{b}}_{i}). 
\end{equation*}
The output of the DNN corresponding to a batch of images is
\begin{equation*}
{\bf{s}}={\bf{W}}_{N}{\bf{h}}_{N-1}+{\bf{b}}_{N}.
\end{equation*}

In this case ${\bf{s}}$ is a $k\times t$ matrix, where $k$ is the number of classes in the classification, and $t$ is the number of training images in the batch. The element ${\bf{s}}_{ij}$ in matrix ${\bf{s}}$ is the score of the $j$-th training image corresponding to the $i$-th class. The total loss of the DNN is calculated as
\begin{equation*}
f \big( \{{\bf{W}}_{1},\dots,{\bf{W}}_{N} \}, \{{\bf{b}}_{1},\dots,{\bf{b}}_{N} \} \big)=-\frac{1}{t}\sum_{j =1}^{t}\mathrm{log}\frac{e^{{\bf{s}}_{y_{j}j}}}{\sum_{i=1}^{k}e^{{\bf{s}}_{ij}}}+ \lambda \sum_{i=1}^{N} \| {\bf{W}}_i \|_{F}^{2},
\end{equation*}
where $ \| \cdot \|_{F}^{2}$ denotes the Frobenius norm, the first term is cross-entropy loss, $y_{j}$ is the correct class of the $j$-th image, and the second term is $L_2$ weight regularization. 

Hereafter, for simplicity of notation we write $ \{{\bf{W}}_{i} \}_{i=1}^N$, or simply $ \{{\bf{W}}_{i} \}$, instead of $ \{{\bf{W}}_{1},\dots,{\bf{W}}_{N} \}$. The same notational convention applies to writing $ \{{\bf{b}}_{i} \}$ instead of $ \{{\bf{b}}_{1},\dots,{\bf{b}}_{N} \}$.
The training of a DNN is a process of minimizing the loss by updating weights and biases. If we use the gradient descent method, the update at every step is
\begin{align*}
{\bf{W}}_{i}&={\bf{W}}_{i}-\alpha \frac{\partial f \big(\{{\bf{W}}_{i}\}, \{{\bf{b}}_{i} \} \big)}{\partial {\bf{W}}_{i}},
\\ {\bf{b}}_{i}&={\bf{b}}_{i}-\alpha \frac{\partial f \big( \{{\bf{W}}_{i} \}, \{{\bf{b}}_{i} \} \big)}{\partial {\bf{b}}_{i}},
\end{align*}
for $i = 1, \ldots, N,$ where $\alpha$ is the learning rate. 

Our objective is to prune the weights of the DNN, and therefore we minimize the loss function subject to constraints on the cardinality of weights in each layer. More specifically, our training process solves
\begin{equation*}
\begin{aligned}
& \underset{ \{{\bf{W}}_{i}\},\{{\bf{b}}_{i} \}}{\text{minimize}}
& & f \big( \{{\bf{W}}_{i} \}, \{{\bf{b}}_{i} \} \big),
\\ & \text{subject to}
& & \mathrm{card}({\bf{W}}_{i})\le l_{i}, \; i = 1, \ldots, N,
\end{aligned}
\end{equation*}
where $\mathrm{card}(\cdot)$ returns the number of nonzero elements of its matrix argument and $l_{i}$ is the desired number of weights in the $i$-th layer of the DNN\footnote{Our framework is also compatible with the constraint of $l$ total number of weights for the whole DNN.}. A prior work \cite{kiaee2016alternating} uses ADMM for DNN training with regularization in the objective function, which can result in sparsity as well. On the other hand, our method directly targets at sparsity with incorporating hard constraints on the weights, thereby resulting in a higher degree of sparsity.

\subsection{Systematic Weight Pruning Framework using ADMM}

We can rewrite the above weight pruning optimization problem as
\begin{equation*}
\begin{aligned}
& \underset{ \{{\bf{W}}_{i}\},\{{\bf{b}}_{i} \}}{\text{minimize}}
& & f \big( \{{\bf{W}}_{i} \}, \{{\bf{b}}_{i} \} \big),
\\ & \text{subject to}
& & {\bf{W}}_{i}\in {\bf{S}}_{i}, \; i = 1, \ldots, N,
\end{aligned}
\end{equation*}
where ${\bf{S}}_{i}= \{{\bf{W}}_{i}\mid \mathrm{card}({\bf{W}}_{i})\le l_{i}  \}, i=1,\dots,N$. It is clear that ${\bf{S}}_{1},\dots,{\bf{S}}_{N}$ are nonconvex sets, and it is in general difficult to solve optimization problems with nonconvex constraints. 
The problem can be equivalently rewritten in a form without constraint, which is
\begin{equation}
\label{admm_ind}
 \underset{ \{{\bf{W}}_{i}\},\{{\bf{b}}_{i} \}}{\text{minimize}}
\ \ \ f \big( \{{\bf{W}}_{i} \}, \{{\bf{b}}_{i} \} \big)+\sum_{i=1}^{N} g_{i}({\bf{W}}_{i}), 
\end{equation}
where $g_{i}(\cdot)$ is the indicator function of ${\bf{S}}_{i}$, i.e.,
\begin{eqnarray*}g_{i}({\bf{W}}_{i})=
\begin{cases}
 0 & \text { if } \mathrm{card}({\bf{W}}_{i})\le l_{i}, \\ 
 +\infty & \text { otherwise. }
\end{cases}
\end{eqnarray*}
The first term of problem (\ref{admm_ind}) is the loss function of a DNN, while the second term is non-differentiable. This problem cannot be solved analytically or by stochastic gradient descent. A recent paper \cite{boyd2011}, however, demonstrates that such problems lend themselves well to the application of ADMM, via a special decomposition into simpler subproblems. We begin by equivalently rewriting the above problem in ADMM form as
\begin{equation*}
\begin{aligned}
& \underset{ \{{\bf{W}}_{i}\},\{{\bf{b}}_{i} \}}{\text{minimize}}
& & f \big( \{{\bf{W}}_{i} \}, \{{\bf{b}}_{i} \} \big)+\sum_{i=1}^{N} g_{i}({\bf{Z}}_{i}),
\\ & \text{subject to}
& & {\bf{W}}_{i}={\bf{Z}}_{i}, \; i = 1, \ldots, N.
\end{aligned}
\end{equation*}
The augmented Lagrangian \cite{boyd2011} of the above optimization problem is given by
\begin{equation*}
\begin{aligned}
 L_{\rho} \big( \{{\bf{W}}_{i} \}, \{{\bf{b}}_{i} \}, \{{\bf{Z}}_{i} \}, \{{\bf{\Lambda}}_{i} \} \big)
 &=f \big( \{{\bf{W}}_{i} \}, \{{\bf{b}}_{i} \} \big)+
 \sum_{i=1}^{N} g_{i}({\bf{Z}}_{i})
 \\&~~+
 \sum_{i=1}^{N} \mathrm{tr} [{\bf{\Lambda}}_{i}^T({\bf{W}}_{i}-{\bf{Z}}_{i}) ]
 +
 \sum_{i=1}^{N} \frac{\rho_{i}}{2} \| {\bf{W}}_{i}-{\bf{Z}}_{i} \|_{F}^{2},
 \end{aligned}
\end{equation*}
where ${\bf{\Lambda}}_{i}$ has the same dimension as ${\bf{W}}_{i}$ and is the Lagrange multiplier (also known as the dual variable) corresponding to the constraint ${\bf{W}}_{i}={\bf{Z}}_{i}$, the positive scalars $ \{\rho_{1},\dots,\rho_{N} \}$ are penalty parameters, $\mathrm{tr}(\cdot)$ denotes the trace, and $ \| \cdot \|_{F}^{2}$ denotes the Frobenius norm. Defining the scaled dual variable ${\bf{U}}_{i}=(1/\rho_{i}){\bf{\Lambda}}_{i}$, the augmented Lagrangian can be equivalently expressed as
\begin{equation*}
\begin{aligned}
L_{\rho} \big( \{{\bf{W}}_{i} \}, \{{\bf{b}}_{i} \},  \{{\bf{Z}}_{i} \},  \{{\bf{\Lambda}}_{i} \} \big)
&=
f \big( \{{\bf{W}}_{i} \}, \{{\bf{b}}_{i} \}\big)+
\sum_{i=1}^{N} g_{i}({\bf{Z}}_{i})
\\&~~+
\sum_{i=1}^{N} \frac{\rho_{i}}{2} \| {\bf{W}}_{i} -{\bf{Z}}_{i}+{\bf{U}}_{i} \|_{F}^{2}
-
\sum_{i=1}^{N} \frac{\rho_{i}}{2} \|{\bf{U}}_{i} \|_{F}^{2}.
\end{aligned}
\end{equation*}
The ADMM algorithm proceeds by repeating, for $k = 0,1, \dots$, the following steps \cite{boyd2011,Liu2013}:
\begin{align}
\label{1}
  \{{\bf{W}}_{i}^{k+1},{\bf{b}}_{i}^{k+1} \}&:=\mathop{\arg\min}_{ \{{\bf{W}}_{i}\},\{{\bf{b}}_{i} \}} \ \ \ L_{\rho} \big( \{{\bf{W}}_{i} \}, \{{\bf{b}}_{i} \}, \{{\bf{Z}}_{i}^{k} \}, \{{\bf{U}}_{i}^{k} \} \big) \\
\label{2}
   \{{\bf{Z}}_{i}^{k+1} \}&:=\mathop{\arg\min}_{ \{{\bf{Z}}_{i} \}} \ \ \ L_{\rho} \big( \{{\bf{W}}_{i}^{k+1} \}, \{{\bf{b}}_{i}^{k+1} \}, \{{\bf{Z}}_{i} \}, \{{\bf{U}}_{i}^{k} \} \big)  \\
\label{3}
  {\bf{U}}_{i}^{k+1}&:={\bf{U}}_{i}^{k}+{\bf{W}}_{i}^{k+1}-{\bf{Z}}_{i}^{k+1},
\end{align}
until both of the following conditions are satisfied
\begin{equation}
\label{eps}
 \| {\bf{W}}_{i}^{k+1}-{\bf{Z}}_{i}^{k+1}  \|_{F}^{2} \ \le \epsilon_{i}, \ \  \| {\bf{Z}}_{i}^{k+1}-{\bf{Z}}_{i}^{k}  \|_{F}^{2} \ \le \epsilon_{i}.
\end{equation}

In order to solve the overall pruning problem, we need to solve subproblems (\ref{1}) and (\ref{2}). More specifically, problem (\ref{1}) can be formulated as
\begin{equation}
\label{4}
 \underset{ \{{\bf{W}}_{i}\},\{{\bf{b}}_{i} \}}{\text{minimize}}
\ \ \ f \big( \{{\bf{W}}_{i} \}, \{{\bf{b}}_{i} \} \big)+\sum_{i=1}^{N} \frac{\rho_{i}}{2}  \| {\bf{W}}_{i}-{\bf{Z}}_{i}^{k}+{\bf{U}}_{i}^{k} \|_{F}^{2}, \\
\end{equation}
where the first term is the loss function of the DNN, and the second term can be considered as a special $L_2$ regularizer. 
Since the regularizer is a differentiable quadratic norm, and the loss function of the DNN is differentiable, problem (\ref{4}) can be solved by stochastic gradient descent.
More specifically, the gradients of the augmented Lagrangian with respect to ${\bf{W}}_{i}$ and ${\bf{b}}_{i}$ are given by
\begin{align*}
\frac{\partial L_{\rho} \big(\{{\bf{W}}_{i} \}, \{{\bf{b}}_{i} \}, \{{\bf{Z}}_{i}^{k} \}, \{{\bf{U}}_{i}^{k} \} \big)}{\partial {\bf{W}}_{i}}&=\frac{\partial f \big(\{{\bf{W}}_{i}\},\{{\bf{b}}_{i}\} \big)}{\partial {\bf{W}}_{i}} + \rho_{i}  ({\bf{W}}_{i}-{\bf{Z}}_{i}^{k}+{\bf{U}}_{i}^{k}), \\
\frac{\partial L_{\rho} \big(\{{\bf{W}}_{i}\},\{{\bf{b}}_{i}\},\{{\bf{Z}}_{i}^{k} \},\{{\bf{U}}_{i}^{k}\} \big)}{\partial {\bf{b}}_{i}}&=\frac{\partial f \big(\{{\bf{W}}_{i}\},\{{\bf{b}}_{i}\} \big)}{\partial {\bf{b}}_{i}}.
\end{align*}
Note that we cannot prove optimality of the solution to subproblem (\ref{1}), just as we can not prove optimality of the solution to the original DNN training problem due to the nonconvexity of the loss function of DNN.

On the other hand, problem (\ref{2}) can be formulated as
\begin{equation*}
 \underset{ \{{\bf{Z}}_{i} \}}{\text{minimize}}
\ \ \ \sum_{i=1}^{N} g_{i}({\bf{Z}}_{i})+\sum_{i=1}^{N} \frac{\rho_{i}}{2} \| {\bf{W}}_{i}^{k+1}-{\bf{Z}}_{i}+{\bf{U}}_{i}^{k} \|_{F}^{2}. \\
\end{equation*}
Since $g_{i}(\cdot)$ is the indicator function of the set ${\bf{S}}_{i}$, the globally optimal solution of this problem can be explicitly derived as \cite{boyd2011}:
\begin{equation}
\label{5}
  {\bf{Z}}_{i}^{k+1} = {{\bf{\Pi}}_{{\bf{S}}_{i}}}({\bf{W}}_{i}^{k+1}+{\bf{U}}_{i}^{k}),
\end{equation}
where ${{\bf{\Pi}}_{{\bf{S}}_{i}}(\cdot)}$ denotes the Euclidean projection onto the set ${\bf{S}}_{i}$. Note that ${\bf{S}}_{i}$ is a nonconvex set, and computing the projection onto a nonconvex set is a difficult problem in general. However, the special structure of ${\bf{S}}_{i}= \{{\bf{W}}\mid \mathrm{card}({\bf{W}})\le l_{i} \}$ allows us to express this Euclidean projection analytically. Namely, the solution of (\ref{2}) is to keep the $l_{i}$ elements of ${\bf{W}}_{i}^{k+1}+{\bf{U}}_{i}^{k}$ with the largest magnitudes and set the rest to zero \cite{boyd2011}. Finally, we update the dual variable ${\bf{U}}_{i}$ according to (\ref{3}). This concludes one iteration of the ADMM algorithm. 

We observe that the proposed systematic framework exhibits multiple major advantages in comparison with the heuristic weight pruning method in \cite{han2015}. Our proposed method achieves a higher compression ratio with a higher convergence rate compared with the iterative pruning and retraining method in \cite{han2015}. For example, we achieve 15$\times$ compression ratio on AlexNet with only 10 iterations of ADMM.
Additionally, subproblem (\ref{1}) can be solved in a fraction of the number of iterations needed for training the original network when we use warm start initialization, i.e., when we initialize subproblem (\ref{1}) with $\{{\bf{W}}_{i}^{k},{\bf{b}}_{i}^{k} \}$ in order to find  $\{{\bf{W}}_{i}^{k+1},{\bf{b}}_{i}^{k+1} \}$. For example, when training on the AlexNet model using the ImageNet data set, convergence is achieved in approximately $\frac{1}{10}$ of the total iterations required for the original DNN training. Also, problems (\ref{2}) and (\ref{3}) are straightforward to carry out, thus their computational time can be ignored. As a synergy of the above effects, the total computational time of 10 iterations of ADMM will be similar to (or at least in the same order of) the training time of the original DNN. Furthermore, we achieve 21$\times$ compression ratio on AlexNet without accuracy loss when we use 40 iterations of ADMM.

\subsection{The Final Retraining Step}

For very small values of $\epsilon_i$ in (\ref{eps}), ADMM needs a large number of iterations to converge. However, in many applications, such as the weight pruning problem considered here, a slight increase in the value of $\epsilon_i$ can result in a significant speedup in convergence.
On the other hand, when ADMM stops early, the weights to be pruned may not be identically zero, in the sense that there will be small nonzero elements contained in ${\bf{W}}_{i}$. To deal with this issue, we keep the $l_{i}$ elements with the largest magnitude in ${\bf{W}}_{i}$, set the rest to zero and no longer involve these elements in training (i.e., we prune these weights). Then, we \emph{retrain the DNN}. Note that we only need a single retraining step and the convergence is much faster than training the original DNN, since the starting point of the retraining is already close to the point which can achieve the original test/validation accuracy.

\subsection{Overall Illustration of Our Proposed Framework}

We take the weight distribution of every (convolutional or fully connected) layer on LeNet-5 as an example to illustrate our systematic weight pruning method. The weight distributions at different stages are shown in Figure 1. The subfigures in the left column show the weight distributions of the pretrained model, which serves as our starting point. The subfigures in the middle column show that after the convergence of ADMM for moderate values of $\epsilon_i$, we observe a clear separation between weights whose values are close to zero and the remaining weights. To prune the weights rigorously, we set the values of the close-to-zero weights exactly to zero and retrain the DNN without updating these values. The subfigures in the right column show the weight distributions after our final retraining step. We observe that most of the weights are zero in every layer. This concludes our weight pruning procedure.

\begin{figure}{}
\centering
\includegraphics[height=9.8cm]{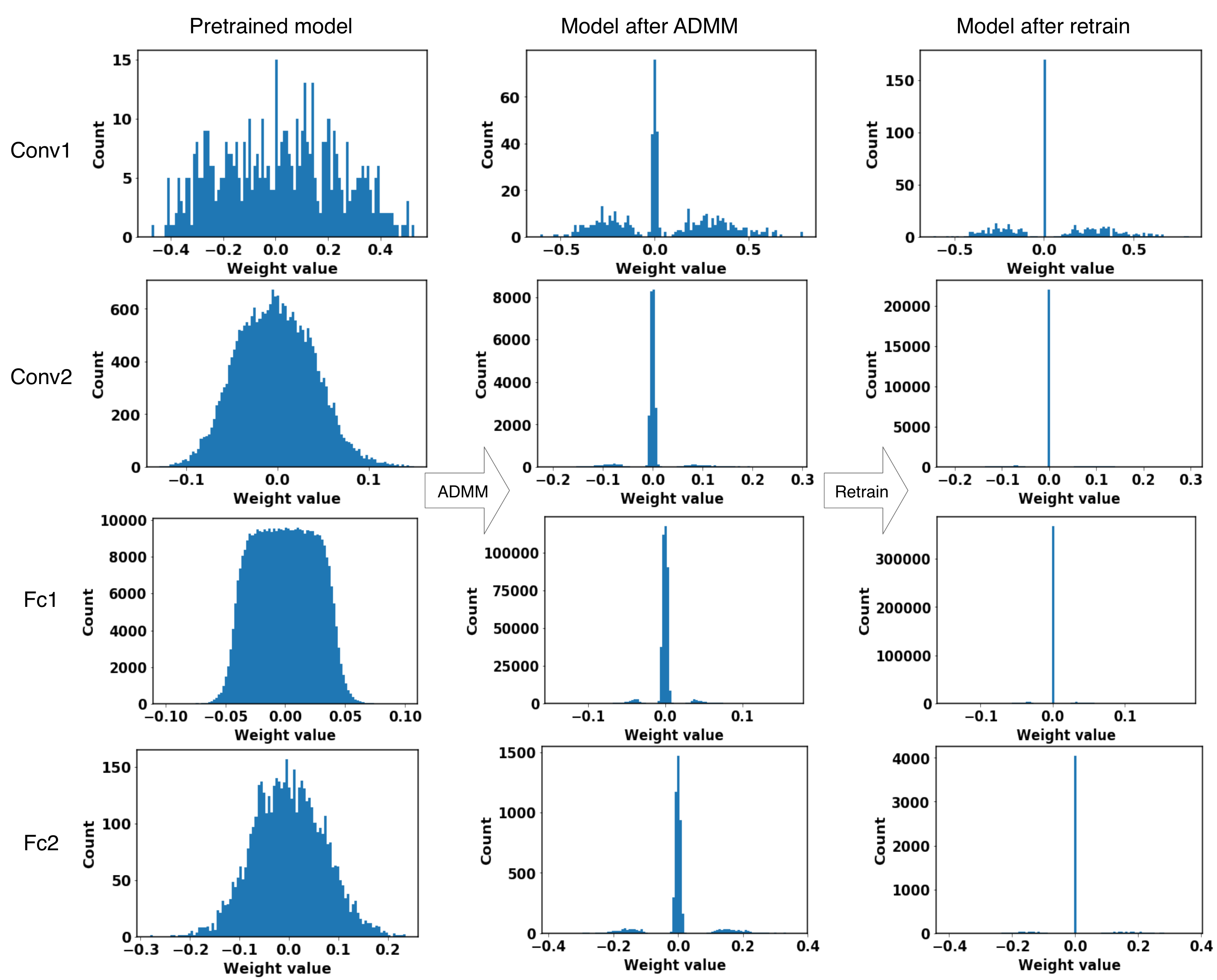}
\caption{Weight distribution of every (convolutional or fully connected) layer on LeNet-5. The subfigures in the left column are the weight distributions of the pretrained DNN model (serving as our starting point); the subfigures of the middle column are the weight distributions after the ADMM procedure; the subfigures of the right column are the weight distributions after our final retraining step. Note that the subfigures in the last column include a small number of nonzero weights that are not clearly visible due to the large number of zero weights.}
\end{figure}

\section{Experimental Results}

\begin{table}{}
\centering
\caption{Weight pruning results on LeNet-300-100 network}
\begin{tabular}{p{1.2cm}p{1.6cm}p{3.2cm}p{3.5cm}p{2.2cm}}
\hline
Layer & Weights & Weights after prune & Weights after prune \% & Result of \cite{han2015} \% \\ \hline
fc1 &  235.2K & 9.41K & 4\% & 8\% \\
fc2 & 30K & 2.1K & 7\% & 9\% \\
fc3 & 1K & 0.12K & 12\% & 26\% \\ \hline
Total & 266.2K & 11.6K & 4.37\% & 8\% \\ \hline
\end{tabular}
\end{table}

\begin{table}
\centering
\caption{Weight pruning results on LeNet-5 network}
\begin{tabular}{p{1.2cm}p{1.6cm}p{3.2cm}p{3.5cm}p{2.2cm}}
\hline
Layer & Weights & Weights after prune & Weights after prune \% & Result of \cite{han2015} \% \\ \hline
conv1 & 0.5K & 0.1K & 20\% & 66\% \\
conv2 & 25K & 2K & 8\% & 12\% \\
fc1 & 400K & 3.6K & 0.9\% & 8\% \\
fc2 & 5K & 0.35K & 7\% & 19\% \\ \hline
Total & 430.5K & 6.05K & 1.4\% & 8\% \\ \hline
\end{tabular}
\end{table}

We have tested the proposed systematic weight pruning framework on the MNIST benchmark using the LeNet-300-100 and LeNet-5 models \cite{lecun1998} and the ImageNet ILSVRC-2012 benchmark on the AlexNet model \cite{krizhevsky2012imagenet}, in order to perform an apple-to-apple comparison with the prior heuristic pruning work \cite{han2015}. The LeNet models are implemented and trained in TensorFlow \cite{Abadi2016} and the AlexNet models are trained in Caffe \cite{jia2014caffe}. We carry out our experiments on NVIDIA Tesla P100 GPUs. The weight pruning results consistently outperform the prior work. On the LeNet-5 model, we achieve 71.2$\times$ weight reduction without accuracy loss, which is 5.9 times compared with \cite{han2015}. On the AlexNet model, we achieve 21$\times$ weight reduction without accuracy loss, which is 2.3 times compared with \cite{han2015}. Moreover, when we focus on the convolutional layer pruning for computation reductions, we can reduce the total computation by five times compared with the prior work \cite{han2015}.

\subsection{Testing results on LeNet models on MNIST data set}

Table 1 shows our per-layer pruning results on the LeNet-300-100 model. LeNet-300-100 is a fully connected network with 300 and 100 neurons on the two hidden layers, respectively, and achieves 98.4$\%$ test accuracy on the MNIST benchmark.
Table 2 shows our per-layer pruning results on the LeNet-5 model. LeNet-5 contains two convolutional layers, two pooling layers and two fully connected layers, and can achieve 99.2$\%$ test accuracy on the MNIST benchmark.

Our pruning framework does not incur accuracy loss and can achieve a much higher compression ratio on these networks compared with the prior iterative pruning heuristic \cite{han2015}, which reduces the number of parameters by 12$\times$ on both LeNet-300-100 and LeNet-5. On the LeNet-300-100 model, our pruning method reduces the number of weights by 22.9$\times$, which is 90\% higher than \cite{han2015}. Also, our pruning method reduces the number of weights by 71.2$\times$ on the LeNet-5 model, which is 5.9 times compared with \cite{han2015}.

\subsection{Testing results on AlexNet model using ImageNet benchmark}

We implement our systematic weight pruning method using the BAIR/BVLC AlexNet model\footnote{\url{https://github.com/BVLC/caffe/tree/master/models/bvlc_alexnet}} on the ImageNet ILSVRC-2012 benchmark. The implementation is on the Caffe tool because it is faster than TensorFlow. The original BAIR/BVLC AlexNet model can achieve a top-5 accuracy 80.2\% on the validation set. AlexNet contains 5 convolutional (and pooling) layers and 3 fully connected layers with a total of 60.9M parameters, with the detailed network structure shown in deploy.prototxt text on the website indicated in footnote 2. 

Our first set of experiments only target model size reductions for AlexNet, and the results are shown in Table 3. It can be observed that our pruning method can reduce the number of weights by 21$\times$ on AlexNet, which is more than twice compared with the prior iterative pruning heuristic. We achieve a top-5 accuracy of 80.2\% on the validation set of ImageNet ILSVRC-2012. Layer-wise comparison results are also shown in Table 3, while comparisons with some other model compression methods are shown in Table 5. These results clearly demonstrate the advantage of the proposed systematic weight pruning framework using ADMM.

\begin{table}
\centering
\caption{Weight pruning results on AlexNet network (purely focusing on weight reductions) without accuracy loss}
\begin{tabular}{p{1.2cm}p{1.6cm}p{3.2cm}p{3.5cm}p{2.2cm}}
\hline
Layer & Weights & Weights after prune & Weights after prune\% & Result of \cite{han2015} \\ \hline
conv1 & 34.8K & 28.19K & 81\% & 84\% \\
conv2 & 307.2K & 61.44K & 20\% & 38\% \\
conv3 & 884.7K & 168.09K & 19\% & 35\% \\
conv4 & 663.5K & 132.7K & 20\% & 37\% \\
conv5 & 442.4K & 88.48K & 20\% & 37\% \\
fc1 & 37.7M & 1.06M & 2.8\% & 9\% \\
fc2 & 16.8M & 0.99M & 5.9\% & 9\% \\
fc3 & 4.1M & 0.38M & 9.3\% & 25\% \\ \hline
Total & 60.9M & 2.9M & 4.76\% & 11\% \\ \hline
\end{tabular}
\end{table}

Our second set of experiments target computation reduction besides weight reduction. Because the major computation in state-of-the-art DNNs is in the convolutional layers, we mainly target weight pruning in these layers. Although on AlexNet, the number of weights in convolutional layers is less than that in fully connected layers, the computation on AlexNet is dominated by its 5 convolutional layers. In our experiments, we conduct experiments which keep the same portion of weights as \cite{han2015} in fully connected layers but prune more weights in convolutional layers. For AlexNet, Table 4 shows that we can reduce the number of weights by 13.4$\times$ in convolutional layers, which is five times compared with 2.7$\times$ in \cite{han2015}. 
This indicates our pruning method can reduce much more computation compared with the prior work \cite{han2015}. Layer-wise comparison results are also shown in Table 4. Still, it is difficult to prune weights in the first convolutional layer because they are needed to directly extract features from the raw inputs. Our major gain is because (i) we can achieve significant weight reduction in conv2 through conv5 layers, and (ii) the first convolutional layer is relatively small and less computational intensive. 

\begin{table}
\centering
\caption{Weight pruning results on AlexNet network (focusing on computation reductions) without accuracy loss}
\begin{tabular}{p{2.4cm}p{1.4cm}p{2.9cm}p{3.1cm}p{1.9cm}}
\hline
Layer & Weights & Weights after prune & Weights after prune\% & Result of \cite{han2015} \\ \hline
conv1 & 34.8K & 21.92K & 63\% & 84\% \\
conv2 & 307.2K & 21.5K & 7\% & 38\% \\
conv3 & 884.7K & 53.08K & 6\% & 35\% \\
conv4 & 663.5K & 46.45K & 7\% & 37\% \\
conv5 & 442.4K & 30.97K & 7\% & 37\% \\
fc1 & 37.7M & 3.39M & 9\% & 9\% \\
fc2 & 16.8M & 1.51M & 9\% & 9\% \\
fc3 & 4.1M & 1.03M & 25\% & 25\% \\ \hline
Total of conv1-5 & 2332.6K & 173.92k & 7.46\% & 37.1\% \\ \hline
\end{tabular}
\end{table}

Several extensions \cite{yang2016,wen2016learning} of the original weight pruning work have improved in various directions such as energy efficiency for hardware implementation and regularity, but they cannot strictly outperform the original work \cite{han2015} in terms of compression ratio under the same accuracy. The very recent work \cite{dai2017} employs an evolutionary algorithm for weight pruning, which incorporates randomness in both pruning and growing of weights following certain probability rules. It can achieve a comparable model size with our work. However, it suffers from a prohibitively long retraining phase. For example, it needs to start with an already-compressed model with 8.4M parameters for further pruning on ImageNet, instead of the original AlexNet model. By using an already-compressed model, it can reduce the number of neurons per layer as well, while such reduction is not considered in our proposed framework.

\begin{table}
\centering
\caption{Weight reduction ratio comparisons using different model compression techniques on the AlexNet model}
\begin{tabular}{p{3.69cm}p{2.1cm}p{1.8cm}p{3.4cm}}
\hline
Network &  Top-5 Error & Parameters  & Weight Reduction Ratio  \\ \hline
Baseline AlexNet \cite{krizhevsky2012imagenet} & 19.8\% & 60.9M & 1.0$\times$ \\ \hline
SVD \cite{denton2014exploiting} & $20.6$\% & 11.9M & 5.1$\times$ \\ \hline
Layer-wise pruning \cite{dong2017learning} & $20.0$\%  & 6.7M & 9.1$\times$ \\ \hline
Network pruning \cite{han2015} & $19.7$\%  & 6.7M & 9.1$\times$ \\ \hline
Our result (10 iterations of ADMM) & $19.8\%$ & 4.06M & 15$\times$ \\ \hline
NeST \cite{dai2017} & $19.7\%$ & 3.9M & 15.7$\times$ \\ \hline
Dynamic surgery \cite{guo2016dynamic} & $20.0\%$ & 3.45M & 17.7$\times$ \\ \hline
Our result (25 iterations of ADMM) & $19.8\%$ & 3.24M & 18.8$\times$ \\ \hline
Our result (40 iterations of ADMM) & $19.8\%$ & 2.9M & 21$\times$ \\ \hline
\end{tabular}
\end{table}

\section{Discussion}

\subsection{Parameters and initialization of ADMM}

For nonconvex problems in general, there is no guarantee that ADMM will converge to an optimal point. ADMM can converge to different points for different choices of initial values $ \{{\bf{Z}}_1^{0},\dots,{\bf{Z}}_N^{0} \}$ and $ \{{\bf{U}}_1^{0},\dots,{\bf{U}}_N^{0} \}$ and penalty parameters $ \{\rho_{1},\dots,\rho_{N} \}$ \cite{boyd2011}. To resolve this limitation, we set the pretrained model $ \{{\bf{W}}_{i}^{p},{\bf{b}}_{i}^{p} \}$, a good solution of $\underset{ \{{\bf{W}}_{i}\},\{{\bf{b}}_{i} \}} {\text{minimize}}\ f \big( \{{\bf{W}}_{i} \}, \{{\bf{b}}_{i} \} \big)$, to be the starting point when we use stochastic gradient descent to solve problem (\ref{4}). We initialize ${\bf{Z}}_i^{0}$ by keeping the $l_{i}$ elements of ${\bf{W}}_i^{p}$ with the largest magnitude and set the rest to be zero. We set ${\bf{U}}_1^{0} =\dots= {\bf{U}}_N^{0} = 0$.
For problem (\ref{4}), if the penalty parameters $ \{\rho_{1},\dots,\rho_{N} \}$ are too small, the solution will be close to the minimum of $f(\cdot)$ but fail to regularize the weights, and the ADMM procedure may converge slowly or not converge at all. If the penalty parameters are too large, the solution may regularize the weights well but fail to minimize $f(\cdot)$, and therefore the accuracy of the DNN will be degradated. In actual experiments, we find that $\rho_{1} = \dots = \rho_{N} = 10^{-4}$ is an appropriate choice for LeNet-5 and LeNet-300-100, and that $\rho_{1} = \dots = \rho_{N} = 1.5 \times 10^{-3}$ works well for AlexNet. 

\subsection{Parameters of the desired number of weights in each layer}

We initialize $l_i$ based on existing results in the literature, then we implement our weight pruning method on the DNN and test its accuracy. If there is no accuracy loss on the DNN, we decrease $l_i$ in every layer proportionally. We use binary search to find the smallest $l_i$ that will not result in accuracy loss.

\subsection{Convergence behavior of ADMM and loss value progression on AlexNet}

Convergence behavior of ADMM (5 CONV layers in AlexNet) is shown in Figure 2 (left sub-figure). The loss value progression of AlexNet is shown in Figure 2 (right sub-figure). We start from an existing DNN model without pruning. After the convergence of ADMM, we remove the weights which are (close to) zero, which results in an increase in the loss value. We then retrain the DNN and the loss decreases to the same level as it was before pruning.
\begin{figure}
\caption{Convergence behavior of ADMM and loss value progression of AlexNet}
\centering
\includegraphics[height=3.9cm]{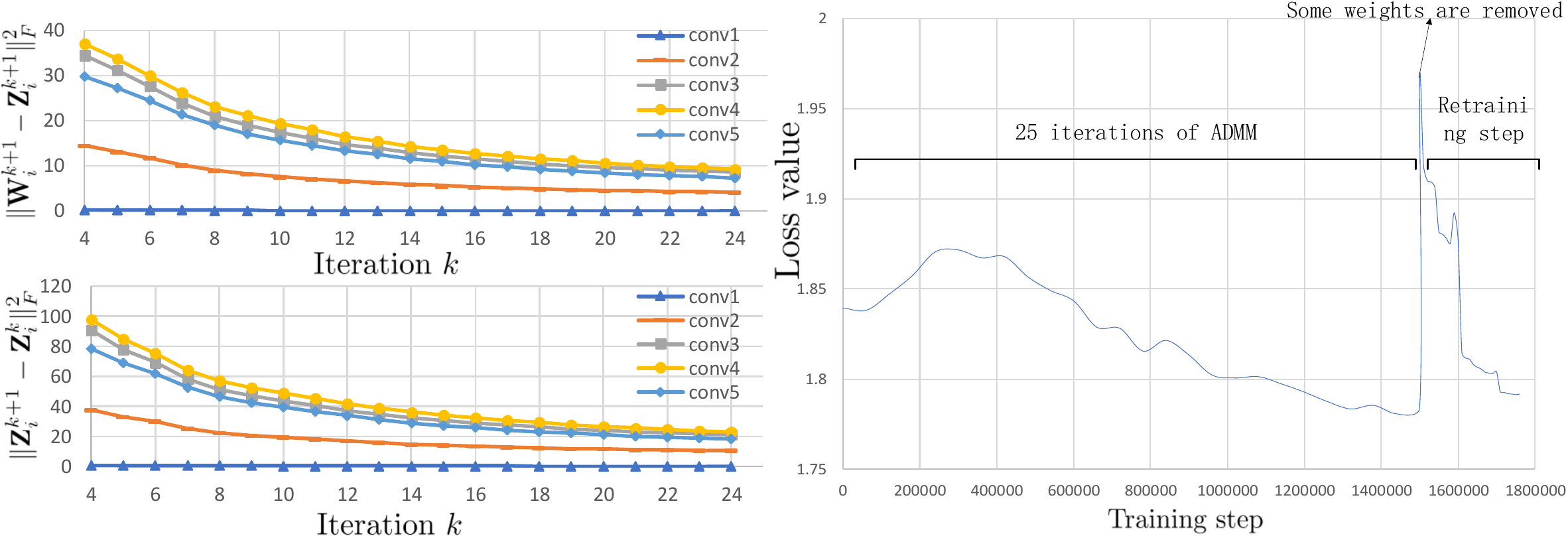}
\end{figure}

\subsection{Discussion on our proposed framework}

The cardinality function is nonconvex and nondifferentiable, which complicates the use of standard gradient algorithms. ADMM circumvents the issue of differentiability systematically, and does so without introducing additional numerical complexity. Furthermore, although ADMM achieves global optimality for convex problems, it has been shown in the optimization literature that it performs extremely well for large classes of nonconvex problems. In fact, ADMM-based pruning can be perceived as a smart regularization technique in which the regularization target will be dynamically updated in each iteration.
The limitation is that we need to tune the parameters $l_i$. However, {\em some} parameter tuning is generally inevitable; even a soft regularization parameter requires fine-tuning in order to achieve the desired solution structure. On the positive side, the freedom in setting $l_i$ allows the user to obtain the exact desired level of sparsity.

\section{Conclusions and Future Work}

In this paper, we presented a systematic DNN weight pruning framework using ADMM. 
We formulate the weight pruning problem of DNNs as a nonconvex optimization problem with combinatorial constraints specifying the sparsity requirements. By using ADMM, the nonconvex optimization problem is decomposed into two subproblems that are solved iteratively, one using stochastic gradient descent and the other analytically. 
We reduced the number of weights by 22.9$\times$ on LeNet-300-100 and 71.2$\times$ on LeNet-5 without accuracy loss. For AlexNet, we reduced the number of weights by 21$\times$ without accuracy loss. When we focued on computation reduction, we reduced the number of weights in convolutional layers by 13.4$\times$ on AlexNet, which is five times compared with the prior work. 

In future work, we will extend the proposed weight pruning method to incorporate structure and regularity in the weight pruning procedure, and develop a unified framework of weight pruning, activation reduction, and weight clustering. 

\subsubsection*{Acknowledgments}

Financial support from the National Science Foundation under
awards CNS-1840813, CNS-1704662 and ECCS-1609916 is gratefully acknowledged.

%
%
%
%

\end{document}